\documentclass[journal]{IEEEtran}

\usepackage{amsmath}
\usepackage{amssymb}
\usepackage{graphicx}
\usepackage{color}
\usepackage{booktabs}
\usepackage{multirow}
\usepackage{cite}
\usepackage{algorithm}
\usepackage{algorithmic}
\usepackage{epstopdf}

\graphicspath{{Figure/}}
\usepackage{fancyhdr}
\makeatletter
\let\ps@headings\ps@fancy
\let\ps@IEEEtitlepagestyle\ps@fancy
\makeatother
\begin{document}

\title{SAGE: Training-Free Semantic Evidence Composition for Edge--Cloud Inference under Hard Uplink Budgets}

\author{Inhyeok~Choi,~\IEEEmembership{}
  and~Hyuncheol Park,~\IEEEmembership{Senior Member,~IEEE}%
\thanks{The authors are with the School of Electrical Engineering,
  Korea Advanced Institute of Science and Technology (KAIST), Daejeon, South Korea.}%
}

\markboth{IEEE Internet of Things Journal}%
{Choi and Park: SAGE for Edge--Cloud Inference under Hard Uplink Budgets}

\maketitle

\begin{abstract}
Edge--cloud hybrid inference offloads difficult inputs to a powerful remote
model, but the uplink channel imposes hard per-request constraints on the
number of bits that can be transmitted.
We show that selecting transmitted content based solely on
attention-based importance, the standard approach in collaborative
inference, is inherently limited under hard budgets.
Two findings support this claim.  First, replacing high-importance units with low-importance but
complementary ones \emph{improves} server accuracy.
This shows that what matters is not individual importance but
how well the transmitted set covers diverse aspects of the input.
Second, spatially uniform selection without any content information
achieves competitive accuracy at moderate budgets.
This confirms that spatial coverage alone carries independent value.  Based on this analysis, we propose \textbf{SAGE} (Semantic
Attention-Guided Evidence), a principled, training-free method that
combines importance filtering with embedding-diversity sampling.
SAGE achieves 93\% of the server ceiling in offloaded accuracy while
transmitting fewer than half of the available evidence units on
ImageNet-1K, substantially outperforming importance-only composition.
\end{abstract}


%% ====================================================================
\section{Introduction}
\label{sec:introduction}
\IEEEPARstart{E}{dge--cloud} hybrid inference systems offload difficult
inputs from a lightweight edge model to a more powerful remote server,
enabling high accuracy without running heavy models on resource-constrained
devices~\cite{zhou2019edge_intelligence, mao2017mec}.  However, the uplink
channel between edge and server is fundamentally resource-limited: bandwidth
caps the payload size, latency requirements bound the transmission duration,
and energy budgets constrain the total communication
cost~\cite{shao2020commcomp, kang2017neurosurgeon}.  In practice, these
heterogeneous physical constraints collectively impose a strict
upper bound on how much data each offloaded request may carry.  We call this
bound a \emph{hard per-request budget}.  Under such a budget,
the edge must select a subset of task-relevant information units, such as
image patches, feature tokens, or compressed representations, to transmit.
We refer to these transmitted units as \emph{semantic evidence}.

The question, then, is not merely \emph{which} evidence to send, but how to
\emph{compose} a limited evidence set that maximizes the server's inference
accuracy.  We refer to this as the \textbf{semantic evidence composition}
problem under hard uplink budgets.

Early solutions relied on split computing, which partitions a DNN at an
intermediate layer and transmits the entire feature
map~\cite{kang2017neurosurgeon, matsubara2023split_survey,
laskaridis2020spinn}, or on learned feature
compression~\cite{shao2020bottlenet, shao2022taskoriented} that reduces the
payload at a fixed split point.  In both cases the transmitted representation
is a monolithic tensor whose size is controlled only by the choice of split
layer or compression ratio; the edge has no ability to select \emph{which
parts} of the input to transmit.

Vision Transformers (ViTs)~\cite{dosovitskiy2021vit, touvron2021deit}
fundamentally change this picture.  Because a ViT decomposes an image into a
sequence of discrete patch tokens, each carrying a localized semantic
feature, the edge can now choose \emph{which tokens} to send---turning
communication reduction into a token selection problem.  Token
pruning~\cite{rao2021dynamicvit, liang2022evit, yin2022avit} and token
merging~\cite{bolya2023tome, bat} have exploited this discrete structure to
reduce computation on a single device, confirming that ViT representations
are highly redundant.  Im~et~al.~\cite{im2024attention} were the first to
leverage this structure for edge--cloud \emph{communication}. The edge DeiT-Tiny's
attention scores rank patches by importance and only the selected subset is
transmitted to a server DeiT-Base, reducing average communication by 68\%.

However, existing approaches in both semantic
communication~\cite{xie2021deepsc, bourtsoulatze2019deepjscc,
gunduz2023beyond} and attention-based
offloading~\cite{im2024attention} optimize for \emph{average}
communication cost rather than guaranteeing per-request feasibility.
Im~et~al.~\cite{im2024attention} select patches by accumulating attention
scores until a threshold~$\theta$ of the total attention mass is reached.
This appears efficient on average because easy inputs have concentrated
attention and require few patches.  However, the inputs actually offloaded
to the server are precisely the hard cases: their attention distributions
are flat and diffuse, so the cumulative criterion retains 140--150 out of
196~patches before reaching the threshold, which may exceed
practical uplink budgets.

Formally, the expected communication cost decomposes as
\begin{equation}
\mathbb{E}[C] =
\Pr(\text{offload})
\cdot
\mathbb{E}[C \mid \text{offload}].
\end{equation}
A low average may simply reflect infrequent offloading while every
offloaded request exceeds the channel capacity.  The metric conflates
offloading frequency with per-request payload size and provides no
deployability guarantee.

We argue that hybrid inference should instead be evaluated using
\emph{deployable collaborative accuracy}: the system accuracy achieved
when every offloaded request must respect a hard uplink budget~$B$.
Under this criterion, the central challenge becomes \emph{semantic evidence
composition}: given only $B$~evidence units (patches), how should the edge
compose an evidence set that best supports the server's decision?

In this work, we show that importance-only evidence composition
is inherently limited under hard budgets and identify
\emph{information coverage} as a key factor that has been overlooked.  When evidence
units are selected purely by importance, the resulting set may
concentrate on a single high-attention region, leaving complementary image areas
underrepresented.  Through server attention analysis and
controlled experiments, we demonstrate that the value of an evidence
unit lies not in its individual importance but in its marginal
contribution to the \emph{coverage} of the transmitted set.

While the importance-diversity trade-off has been studied for
computational token reduction~\cite{bat, alvar2025divprune,
baek2026agile}, no existing work in collaborative
inference~\cite{im2024attention, sana2024adaptive, park2024vit_iaq}
considers information coverage among transmitted units.  We bridge this
gap by demonstrating that coverage-aware composition yields substantial
accuracy gains under hard uplink budgets.

Based on this analysis, we propose \textbf{SAGE} (Semantic
Attention-Guided Evidence), a principled, training-free method that
combines attention-based importance filtering with embedding
farthest-point sampling (FPS) to maximize information coverage within
the budget.

The main contributions are:
\begin{itemize}
\item We formalize the \emph{semantic evidence composition} problem under
  hard uplink budgets and propose \emph{deployable collaborative accuracy}
  as the evaluation criterion.
\item We demonstrate that importance-only composition
  is inherently limited: server attention
  analysis reveals that the value of an evidence unit lies in its marginal
  contribution to \emph{information coverage}, not in its
  individual importance.
\item We show that \emph{spatial coverage carries independent value} in
  the communication setting, bridging the importance-diversity
  literature~\cite{bat, alvar2025divprune, baek2026agile} with
  collaborative inference.
\item We propose SAGE, a principled, training-free method
  requiring no model modifications, and validate its robustness across
  budgets, operating points, and backbone architectures on ImageNet-1K.
\end{itemize}

%% ====================================================================
\section{Related Work}
\label{sec:related}

\subsection{Semantic Communication and Split Inference}

Semantic communication has emerged as a paradigm shift from bit-level
fidelity to meaning-level efficiency.  Xie~et~al.~\cite{xie2021deepsc}
proposed DeepSC, the first Transformer-based end-to-end semantic
communication system for text.  For images, Bourtsoulatze~et~al.
\cite{bourtsoulatze2019deepjscc} introduced DeepJSCC, which directly maps
pixels to channel symbols via learned joint source-channel coding, avoiding
the cliff effect of separate source and channel codes.  Kurka and
G\"{u}nd\"{u}z~\cite{kurka2021deepjsccl} extended this to bandwidth-agile
transmission that adapts to variable channel conditions.
Gunduz~et~al.~\cite{gunduz2023beyond} provided a comprehensive tutorial on
context-aware, semantic, and task-oriented communications, arguing that
transmitted data should be tailored to the downstream task rather than
reconstructed faithfully.  Yang~et~al.~\cite{yang2023semcom_survey}
classified semantic communication into semantic-oriented, goal-oriented, and
semantic-aware categories, while Chaccour~et~al.
\cite{chaccour2025lessdata} advocated for reasoning-driven AI-native
designs.  Strinati and Barbarossa~\cite{strinati20216g} articulated the
vision of effectiveness-level communication for 6G networks.

Split computing partitions a DNN between edge and cloud to balance
computation and communication.  Kang~et~al.~\cite{kang2017neurosurgeon}
pioneered layer-granularity DNN partitioning with latency and energy
prediction, while Laskaridis~et~al.~\cite{laskaridis2020spinn} combined
early-exit with split inference to adapt to dynamic network conditions.
Matsubara~et~al.~\cite{matsubara2023split_survey} provided a comprehensive
survey covering head-network distillation, bottleneck injection, and
supervised compression for split DNNs.
Traditional fronthaul compression focuses on quantizing raw baseband
signals under capacity constraints~\cite{heo2017fronthaul}, whereas
task-oriented approaches shift the design objective from bit-level
fidelity to downstream task performance.
Feature compression is a key enabler of split inference.
Shao and Zhang~\cite{shao2020bottlenet} proposed BottleNet++ for end-to-end
feature compression in device-edge co-inference, achieving up to $256\times$
compression with minimal accuracy loss.  The same authors
\cite{shao2020commcomp} formalized the communication--computation trade-off,
and later introduced an information bottleneck (IB)
approach~\cite{shao2022taskoriented} for jointly optimizing feature
extraction and encoding in a task-oriented manner.  Zhou~et~al.
\cite{zhou2019edge_intelligence} and Mao~et~al.~\cite{mao2017mec}
established the broader edge intelligence and mobile edge computing
frameworks within which these approaches operate.

Our work falls under the task-oriented branch of semantic communication:
the goal is accurate classification, and the communication budget is a hard
constraint.  Unlike DeepJSCC, which learns an end-to-end encoder--decoder,
or split-computing approaches that compress intermediate feature maps,
we use frozen pretrained Vision Transformers and select which input patches
to transmit under a hard evidence budget, exploiting the discrete token
structure of ViTs without learned feature compression.

\subsection{Efficient Vision Transformers}

Since Vision Transformers (ViT)~\cite{dosovitskiy2021vit} process images as
sequences of patch tokens, reducing the number of tokens directly reduces
computation.  Touvron~et~al.~\cite{touvron2021deit} showed that
data-efficient training and distillation yield competitive ViTs (DeiT) even
without large-scale pretraining.

\textbf{Token pruning} methods learn to discard redundant tokens.
DynamicViT~\cite{rao2021dynamicvit} inserts lightweight prediction modules
to hierarchically prune tokens, reducing FLOPs by 31--37\%.
EViT~\cite{liang2022evit} identifies attentive tokens via CLS attention and
fuses inattentive tokens into a single representative.
A-ViT~\cite{yin2022avit} adapts the number of active tokens per layer and
per image using adaptive computation time.  ATS~\cite{fayyaz2022ats}
provides a differentiable, parameter-free sampling module that produces
input-adaptive token counts.

\textbf{Token merging} approaches combine similar tokens instead of
discarding them.  ToMe~\cite{bolya2023tome} uses bipartite soft matching to
merge tokens in a training-free manner, achieving up to $2\times$ throughput with
negligible accuracy loss.  Beyond Attentive Tokens
(BAT)~\cite{bat} jointly optimizes token importance and diversity,
demonstrating that diversity among retained tokens is as important as
individual importance.

\textbf{Importance-diversity trade-off and coverage.}
Recent works have recognized that importance-only token selection leads
to \emph{duplicative redundancy} among retained
tokens~\cite{alvar2025divprune}.
DivPrune~\cite{alvar2025divprune} formulated token pruning as a max-min
diversity problem (MMDP), showing that importance-based retention increases
redundancy and that maximizing the minimum pairwise distance among retained
tokens restores lost information.
AgilePruner~\cite{baek2026agile} conducted a systematic empirical study of
the attention-vs-diversity trade-off and found that attention-based methods
work well on simple images with concentrated features, while
diversity-based methods are superior on complex images whose features are
spatially distributed---precisely the hard cases that are offloaded to a
remote server in our setting.
Beyond the diversity framing, Zheng~et~al.~\cite{zheng2023ccs} provided
a complementary \emph{coverage} perspective in the data pruning setting:
at high pruning rates, importance-only coreset selection performs worse
than random sampling because it leaves \emph{coverage gaps}---regions of
the data distribution that are entirely unrepresented.  Their coverage-
centric selection (CCS) extends the classical set cover problem to
distributions and achieves up to $19$\,pp higher accuracy than
importance-only methods at 90\% pruning on CIFAR-10.  While CCS addresses
dataset-level pruning, the principle directly applies to our per-image
evidence composition: under hard budgets, importance-only selection
leaves coverage gaps in the image, degrading server accuracy.

All of the above methods target computational efficiency on a single
device or dataset-level sample selection.
We repurpose their insights for a fundamentally different goal:
\emph{communication efficiency} in edge--cloud collaborative inference
under hard uplink budgets.  Crucially, the coverage problem is
\emph{amplified} in the communication setting: when tokens are pruned for
computation, the remaining tokens still share the same model context;
when tokens are selected for \emph{transmission}, the server receives
\emph{only} the selected subset with no access to the discarded tokens.
Coverage gaps among transmitted tokens therefore cause a direct information
loss that cannot be recovered.  This distinction motivates our focus on
coverage-aware evidence \emph{composition} rather than token
\emph{reduction}.

\subsection{Attention-Aware Collaborative Inference}

The most directly related line of work combines ViT attention with
collaborative inference.  Im~et~al.~\cite{im2024attention} proposed
attention-aware patch selection where the edge DeiT-Tiny's mean attention
scores determine which patches are sent to the server DeiT-Base, reducing
communication by 68\%.  However, this approach uses a simple top-$k$
attention ranking and optimizes average communication cost rather than
guaranteeing per-request budget feasibility.

Recent works have begun addressing adaptive token selection for
communication.  Devoto~et~al.~\cite{sana2024adaptive} proposed a trainable
token selection mechanism within a JSCC encoder for goal-oriented
communication, dynamically selecting tokens per input under user-specified
rates.  Park~et~al.~\cite{park2024vit_iaq} leveraged ViT attention to
assign variable quantization bit-widths per patch, optimizing bit allocation
rather than patch selection.

A common feature of all existing collaborative inference methods is that
evidence selection is \emph{purely importance-driven}: tokens or patches
are ranked by a scalar importance score (attention, gradient, or learned
gate), and the top-ranked units are transmitted.  None of these works
consider the \emph{diversity} among transmitted units or analyze the
redundancy that importance-only selection introduces.  As the efficient
ViT literature has shown~\cite{bat, alvar2025divprune, baek2026agile},
importance-only retention leads to duplicative redundancy among selected
tokens.  This problem is especially severe in the communication setting,
where the server has \emph{no access} to discarded tokens and therefore
cannot recover from redundant transmissions.

Our work differs from these approaches in two key aspects: (1)~we impose a
\emph{hard per-request evidence budget} rather than optimizing average rate,
and (2)~we formalize the \emph{evidence composition} problem, which
jointly optimizes importance and coverage among transmitted evidence
units, bridging the gap between the computational diversity literature
and communication-constrained collaborative inference.

%% ====================================================================
\section{Problem Analysis}
\label{sec:problem}

\subsection{System Model}

We consider a local-first hybrid inference system in which a lightweight
edge model and a powerful remote server model collaborate under uplink
resource constraints.

Given an input~$x$, the edge model produces a local prediction
$p_{\text{local}}(y|x)$ and extracts $N$~semantic evidence units
$\mathcal{E} = \{e_1, \dots, e_N\}$.  In a ViT-based system, these
correspond to patch tokens ($N\!=\!196$ for a $14\!\times\!14$ grid), but
the formulation applies to any discrete evidence structure.

A confidence gate $g(x)$ determines whether the input is resolved locally
($g(x)\!=\!0$) or offloaded to the server ($g(x)\!=\!1$).  Only offloaded
inputs incur uplink communication.

The uplink channel imposes physical constraints (bandwidth, latency,
energy) that collectively limit the amount of evidence transmittable per
request.  We abstract these into a single \emph{hard budget}~$B$: the
maximum number of evidence units that can be transmitted.  Given a fixed evidence unit size (e.g., a $16\!\times\!16$
image patch), $B$ determines the total payload size.  Given channel
conditions, this payload in turn dictates the transmission latency and
energy cost.

For offloaded inputs, a \emph{composer} $C(\cdot)$ selects a subset
\begin{equation}
C(\mathcal{E}, B) \subseteq \mathcal{E},
\quad
|C(\mathcal{E}, B)| \le B
\end{equation}
which is transmitted to the server.  The server produces a prediction
\begin{equation}
p_{\text{server}}(y \mid C(\mathcal{E}, B)).
\end{equation}
The deployed system returns the local prediction when $g(x)\!=\!0$ and the
server prediction when $g(x)\!=\!1$.

We define \textbf{deployable collaborative accuracy} as the overall system
accuracy of this budget-feasible pipeline.  Unlike average-cost metrics,
deployable accuracy requires that \emph{every} offloaded request satisfies
the uplink budget.
We evaluate performance with two complementary metrics.
\emph{Offloaded accuracy} measures classification accuracy on images
escalated to the server, directly reflecting the quality of evidence
composition.  \emph{Overall accuracy} measures system-level accuracy
including both locally-resolved and offloaded images, capturing the
end-to-end impact.

\subsection{Evidence Composition under Hard Budgets}

The central question is: given only $B$~evidence units, how should the edge
\emph{compose} the transmitted set to maximize server accuracy?  We call
this the \textbf{semantic evidence composition} problem.

A na\"{i}ve approach ranks evidence units by individual importance (e.g.,
attention score) and selects the top~$B$.  We show empirically that this
produces compositions with poor information coverage: high-importance
units cluster in the same semantic region, wasting budget on overlapping
information.  Effective composition requires optimizing both the
\emph{importance} and the \emph{coverage} of selected units jointly.

\begin{figure}[t]
\centering
\includegraphics[width=\linewidth]{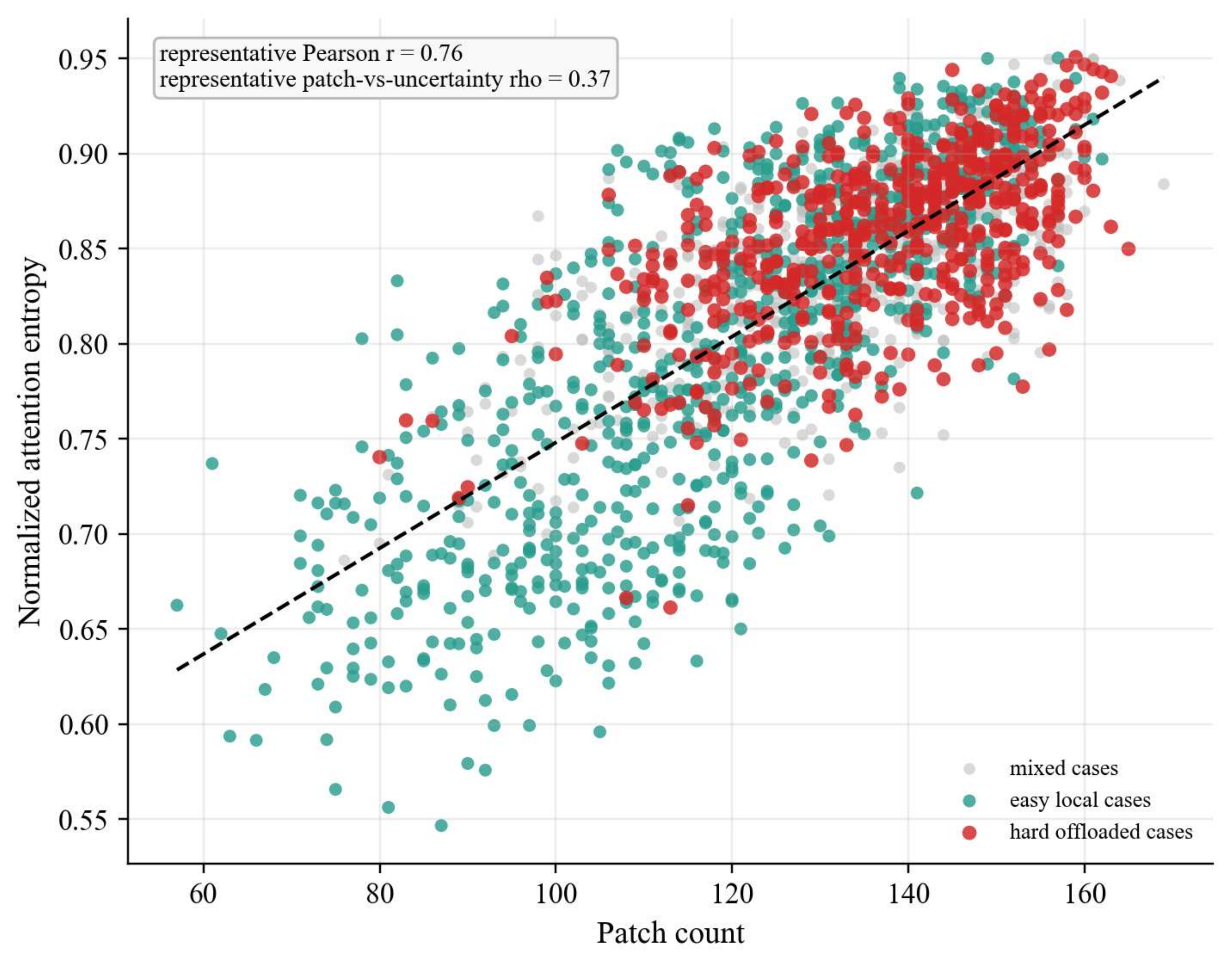}
\caption{Patch count vs.\ normalized attention entropy.  Hard offloaded
cases (red) require more patches and exhibit flatter attention than easy
local cases (green), demonstrating that offloaded inputs are
systematically more demanding.  Pearson $r = 0.76$.}
\label{fig:patch_attention}
\end{figure}

Fig.~\ref{fig:patch_attention} provides empirical motivation for the
hard-budget formulation.  Under a representative offloading setting, images
resolved locally at the edge require few patches and exhibit concentrated
attention distributions (low normalized attention entropy), while images
offloaded to the server demand substantially more patches and exhibit
flatter, more diffuse attention.  The Pearson correlation between
patch count and attention entropy is $r = 0.76$.

This correlation has a critical implication for deployment.  Under an
average-cost framework, the system may appear efficient because easy inputs
(which dominate) require little communication.  However, an
average-cost metric can obscure per-request infeasibility: it provides no
guarantee that any individual offloaded request will fit within the uplink
constraint.

\begin{figure}[t]
\centering
\includegraphics[width=\linewidth]{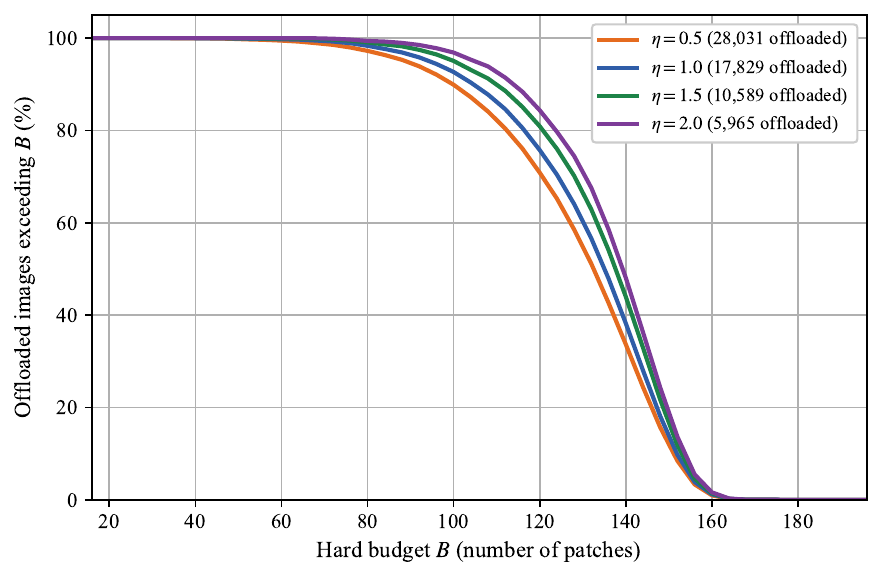}
\caption{Fraction of offloaded images whose patch count under
cumulative attention thresholding ($\theta\!=\!0.95$) exceeds the hard
budget~$B$, on ImageNet-1K.  At $B\!=\!64$, over 99\% of offloaded
images are budget-infeasible across all~$\eta$.}
\label{fig:budget_exceed}
\end{figure}

Fig.~\ref{fig:budget_exceed} quantifies this gap.  We measure the
fraction of offloaded images whose patch count under the standard
cumulative attention threshold ($\theta\!=\!0.95$)~\cite{im2024attention}
exceeds a given budget~$B$.  The results are striking: even when allowing one-third of all
patches ($B\!=\!64$), over 99\% of offloaded images still exceed the
budget regardless of~$\eta$.  At the more permissive $B\!=\!96$, which
permits nearly half of all patches, more than 94\% still exceed it
under $\eta\!=\!1.0$.  The
threshold criterion was designed for average-cost optimization, a
legitimate design goal, but these numbers show that it provides
essentially no per-request feasibility guarantee for the inputs that
actually reach the server.

By contrast, the hard-budget formulation requires the selector to produce a
budget-feasible evidence set for \emph{every} offloaded request, forcing
the system to make principled allocation decisions even for the most
difficult inputs.

\subsection{Why Importance Alone Is Not Enough}
\label{sec:coverage_gap}

We now analyze why importance-only selection
underperforms under hard budgets.  We consider the attention-prefix
baseline~\cite{im2024attention}, which selects the top-$B$ evidence units
by descending attention score.

In a ViT, each patch token encodes a local semantic feature via its
embedding vector~$\mathbf{z}_i$.  Because high-attention patches
concentrate on the most salient object region, importance-only selection
produces evidence sets that cluster around a single area while leaving
complementary image regions underrepresented.
The problem is not merely that
selected units have high pairwise embedding similarity, but that the
evidence set fails to \emph{cover} the diverse aspects of the image that
the server needs for accurate classification.

\textbf{Evidence 1: Individual importance does not determine value.}
To test whether importance-only selection fails because it misses
individually important patches, we compare the patch sets selected by
Attention Prefix and SAGE (proposed in
Section~\ref{sec:method}).  For each image, we classify patches into
three groups: \emph{shared} (selected by both), \emph{added} (selected
only by SAGE), and \emph{dropped} (selected only by Attention Prefix).
We then measure the server model's CLS attention weight for each group
using a full-information ($N\!=\!196$) forward pass.

Table~\ref{tab:server_utility} shows that the patches
SAGE adds have $3\times$ \emph{lower} server attention than the patches
it drops.  In fewer than 25\% of images do the added patches receive
higher server attention than the dropped ones.  Yet SAGE consistently
improves accuracy by $+2$--$4$\,pp.
This reveals that the value of an evidence unit is not its individual
importance, as measured by either client or server attention, but its
\emph{marginal contribution to the information coverage} of the
transmitted set.  Low-importance patches that provide complementary
information are more valuable than high-importance patches that
duplicate already-covered content.

\begin{table}[t]
\centering
\caption{Server attention analysis of swapped patches.
SAGE adds patches with lower server attention than the ones it drops,
yet improves accuracy.  ``Better~\%'' = fraction of images where
added patches have higher server attention than dropped patches.}
\label{tab:server_utility}
\small
\setlength{\tabcolsep}{4pt}
\begin{tabular}{c|c|cc|c|c}
\hline
$B$ & Swap & \multicolumn{2}{c|}{Server Attention} & Better & $\Delta$Acc \\
    & ratio & Added & Dropped & \% & (pp) \\
\hline
32 & 25.8\% & 0.0028 & 0.0096 & 19.2 & +2.3 \\
48 & 20.0\% & 0.0027 & 0.0075 & 23.5 & +4.4 \\
64 & 16.8\% & 0.0027 & 0.0064 & 28.3 & +2.9 \\
96 & 13.2\% & 0.0028 & 0.0059 & 32.6 & +1.7 \\
\hline
\end{tabular}
\end{table}

\begin{figure}[t]
\centering
\includegraphics[width=\linewidth]{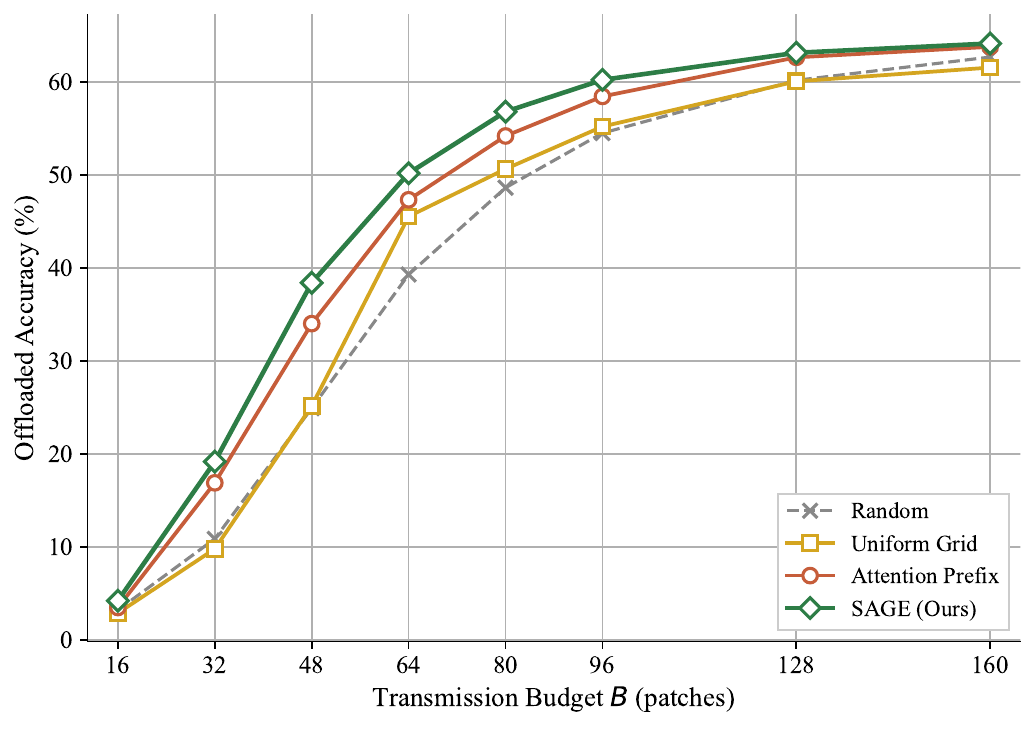}
\caption{Offloaded accuracy vs.\ budget on ImageNet-1K ($\eta\!=\!1.0$).
Random and Uniform Grid use no attention information; Attention
Prefix uses importance only; SAGE combines importance with coverage.
Uniform Grid surpasses Random by over $+6$\,pp at $B\!=\!64$, confirming
that spatial coverage carries independent value.  SAGE consistently
achieves the highest accuracy by combining both importance and coverage.}
\label{fig:4methods}
\end{figure}

\textbf{Evidence 2: Coverage has independent value.}
To isolate the contribution of spatial coverage, we compare four
selection strategies: Random (no importance, no coverage), Uniform Grid
(no importance, coverage only), Attention Prefix (importance only,
no coverage), and SAGE (importance + coverage).
Fig.~\ref{fig:4methods} reports full results.  The key observation is that Uniform Grid, which selects
patches at fixed spatial positions without examining image content,
outperforms Random selection by over $6$\,pp at $B\!=\!64$ and achieves
96\% of Attention Prefix's accuracy.  This confirms that \emph{spatial
coverage carries independent value} under hard budgets: even without
any importance information, ensuring that the evidence set covers diverse
image regions substantially improves server accuracy.

\subsection{Coverage as the Missing Ingredient}
\label{sec:coverage}

The two findings above establish that effective evidence composition
requires not only \emph{importance} (selecting task-relevant units) but
also \emph{information coverage} (ensuring the selected set spans
complementary aspects of the image).  Under generous budgets, coverage
gaps are tolerable because the budget is large enough to cover the image
even with overlap.  Under hard budgets, every unit must count, making
coverage essential.
While the importance-diversity trade-off has been studied for
computational token reduction~\cite{bat, alvar2025divprune,
baek2026agile}, those works address efficiency within a single model,
where the forward pass retains context from discarded tokens.  In the
communication setting, the server receives \emph{only} the transmitted
subset: information in discarded evidence units is permanently lost.
Coverage gaps therefore impose a strictly harder penalty, making
coverage-aware composition not merely beneficial but \emph{essential}.

This motivates a simple design principle: \emph{filter by importance
first, then maximize coverage within the filtered set}.  We instantiate
this principle in the following section.

\begin{figure}[t]
\centering
\includegraphics[width=\linewidth]{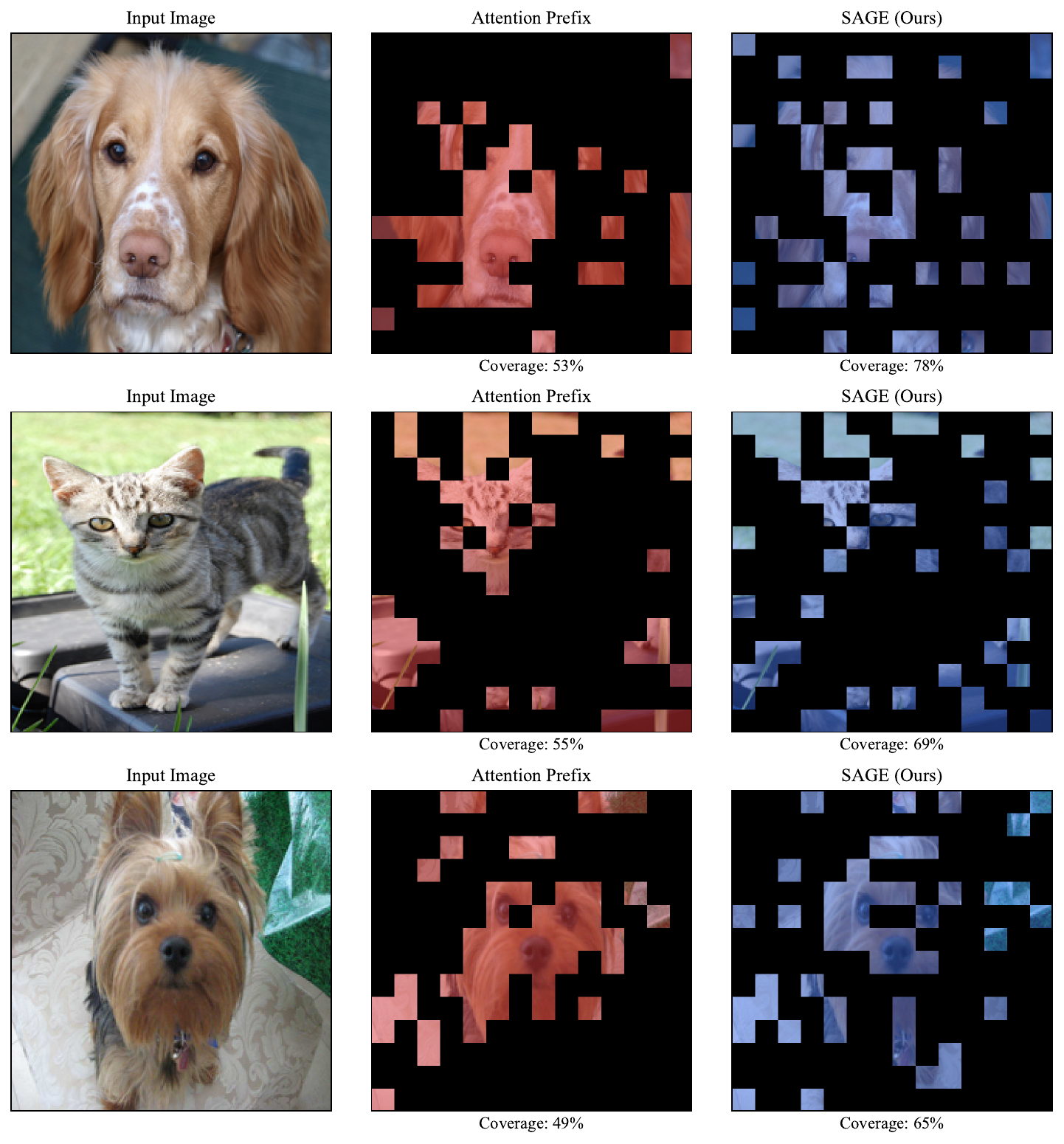}
\caption{Qualitative comparison of patch selection on ImageNet-1K
($B\!=\!48$).  Selected patches are shown in color; unselected patches
are masked in black.  Attention Prefix (red) clusters selections around
the most salient region, while SAGE (blue) distributes them across
complementary areas.  Coverage percentages indicate $7\!\times\!7$
coarse-grid coverage; aggregate statistics are reported in
Table~\ref{tab:coverage_stats}.}
\label{fig:sage_qualitative}
\end{figure}

\begin{figure*}[t]
\centering
\includegraphics[width=0.85\linewidth]{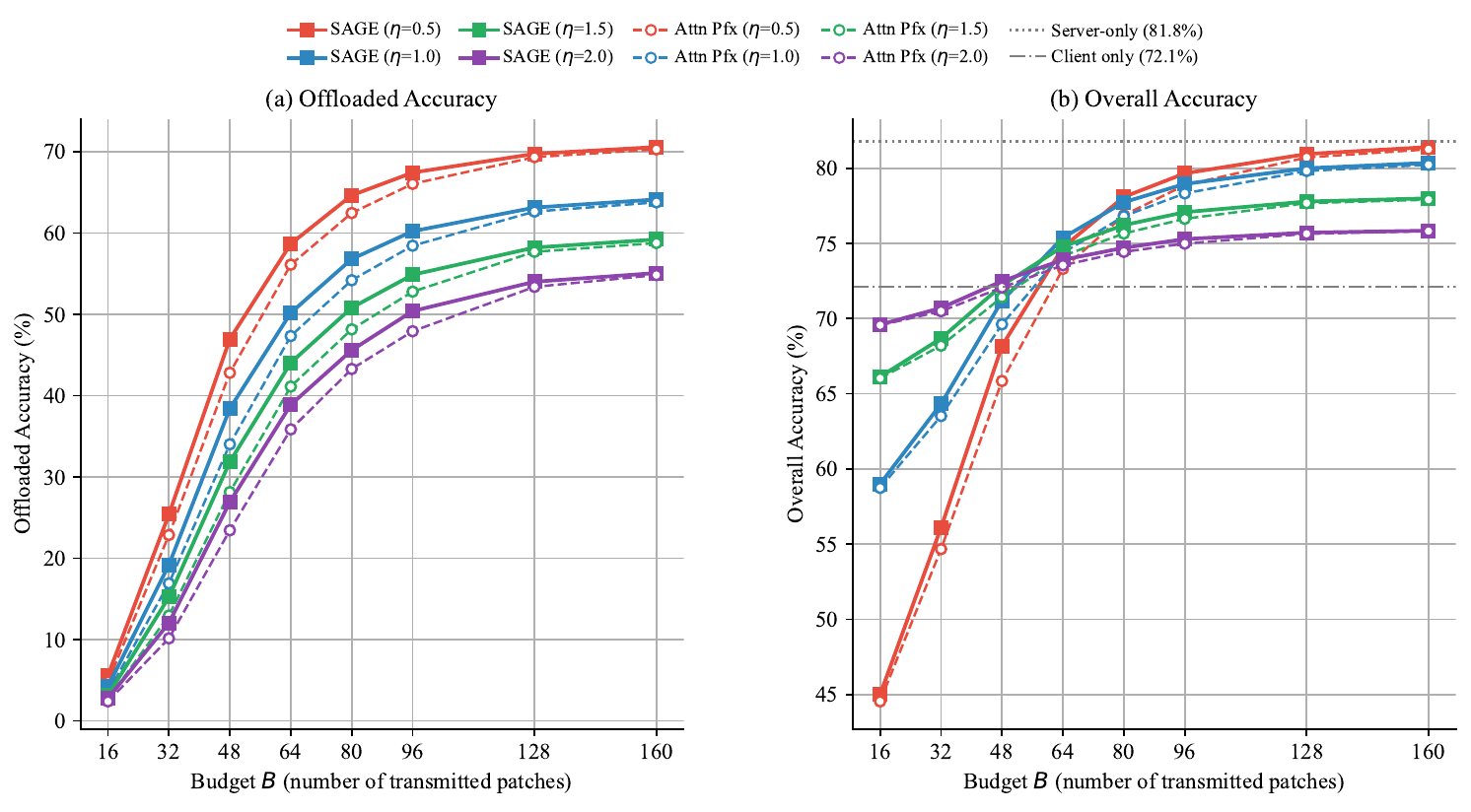}
\caption{Budget--accuracy trade-off on ImageNet-1K under four
confidence-gate settings ($\eta \in \{0.5, 1.0, 1.5, 2.0\}$).
(a)~Offloaded accuracy and (b)~overall accuracy.  Solid lines: SAGE;
dashed lines: Attention Prefix.  Horizontal dashed lines mark the server-only accuracy
(81.8\%) and client-only accuracy (72.1\%).  SAGE consistently outperforms Attention
Prefix across all budgets and operating points, with the largest gains at
low budgets where coverage gaps are most severe.}
\label{fig:budget_curve}
\end{figure*}

%% ====================================================================
\section{SAGE: Proposed Method}
\label{sec:method}

Motivated by the coverage analysis in Section~\ref{sec:coverage}, we
propose \textbf{SAGE}, a principled, training-free method that
instantiates the importance-then-coverage principle.  SAGE requires no
additional parameters or model modifications, yet demonstrates that
coverage-aware evidence composition yields substantial gains over
importance-only selection.

SAGE decomposes evidence composition into two stages: \emph{importance
filtering} (retain only task-relevant candidates) followed by
\emph{coverage maximization} (select the most complementary subset).

\textbf{Intuition.}
Among patches that are sufficiently important (high attention), the system
should prefer those that are semantically \emph{different} from
already-selected patches.  This ensures that the limited budget covers
distinct aspects of the image rather than repeatedly sampling the same
region.  Fig.~\ref{fig:sage_qualitative} illustrates this contrast:
Attention Prefix concentrates selections on a single salient region, while
SAGE distributes them across complementary areas of the image.

\textbf{Algorithm.}
Given the edge model's mean attention vector $\mathbf{a} \in \mathbb{R}^N$
and patch embeddings $\mathbf{Z} \in \mathbb{R}^{N \times D}$, SAGE
operates as follows (Algorithm~\ref{alg:sage}):

\begin{enumerate}
\item \textbf{Attention prefilter}: Retain the top-$2B$ patches by
  attention score as the candidate set~$\mathcal{C}$.  This ensures all
  candidates are at least moderately important while providing enough
  room for diversity-based selection.
\item \textbf{Seed}: Select the highest-attention candidate as the first
  patch: $s_1 = \arg\max_{i \in \mathcal{C}} a_i$.
\item \textbf{Iterative FPS}: For each subsequent selection, compute the
  maximum cosine similarity between each remaining candidate and all
  already-selected patches.  Select the candidate with the
  \emph{lowest} maximum similarity (i.e., the most diverse):
  \begin{equation}
  s_t = \arg\min_{i \in \mathcal{C} \setminus S_{t-1}}\;
  \max_{j \in S_{t-1}}\,
  \hat{\mathbf{z}}_i^{\top} \hat{\mathbf{z}}_j
  \label{eq:sage_fps}
  \end{equation}
  where $\hat{\mathbf{z}}_i = \mathbf{z}_i / \|\mathbf{z}_i\|$ is the
  unit-norm embedding and $S_{t-1}$ is the set of previously selected
  patches.
\item Repeat until $|S| = B$.
\end{enumerate}

\textbf{Key design choice.}
After the prefilter, SAGE does \emph{not} use attention scores in the
iterative selection---only embedding diversity drives the remaining
choices.  The rationale is that the prefilter already guarantees minimum
importance; within this filtered set, maximizing information coverage
yields better server accuracy than re-weighting by attention,
as confirmed by the server attention analysis in
Table~\ref{tab:server_utility}.

\begin{algorithm}[t]
\caption{SAGE: Semantic Attention-Guided Evidence}
\label{alg:sage}
\begin{algorithmic}[1]
\REQUIRE Attention $\mathbf{a} \in \mathbb{R}^N$, embeddings
$\mathbf{Z} \in \mathbb{R}^{N \times D}$, budget $B$
\ENSURE Selected set $S$ with $|S| = B$
\STATE $\mathcal{C} \leftarrow \text{top-}2B\text{ indices by } \mathbf{a}$
\COMMENT{Attention prefilter}
\STATE $\hat{\mathbf{Z}} \leftarrow \mathbf{Z}[\mathcal{C}]\,/\,
\|\mathbf{Z}[\mathcal{C}]\|_{\text{row}}$
\COMMENT{$\ell_2$-normalize}
\STATE $S \leftarrow \{\arg\max_{i \in \mathcal{C}} a_i\}$
\COMMENT{Seed with highest attention}
\FOR{$t = 2, \dots, B$}
  \STATE $\text{sim}_i \leftarrow \max_{j \in S}\;
  \hat{\mathbf{z}}_i^{\top} \hat{\mathbf{z}}_j,
  \quad \forall\, i \in \mathcal{C} \setminus S$
  \STATE $S \leftarrow S \cup
  \{\arg\min_{i \in \mathcal{C} \setminus S}\; \text{sim}_i\}$
  \COMMENT{Most diverse}
\ENDFOR
\RETURN $S$
\end{algorithmic}
\end{algorithm}

%% ====================================================================
\section{Experiments}
\label{sec:experiments}

\subsection{Setup}

\textbf{Models.}
The edge model is DeiT-Tiny (patch size~16, $224\!\times\!224$ input) and the
server model is DeiT-Base, both pretrained on ImageNet-1K.  Each image
produces $N\!=\!196$ candidate patches ($14\!\times\!14$ grid).
Both models are frozen; no fine-tuning is performed.
When receiving all 196 patches, the server achieves 64.4\%
offloaded accuracy and the overall system reaches 80.4\% accuracy
(under $\eta\!=\!1.0$); we refer to these as the \emph{server ceiling}.

\textbf{Dataset.}
We evaluate on \emph{ImageNet-1K} (50{,}000 validation images,
1{,}000 classes).  All 50{,}000 images are used; under $\eta\!=\!1.0$,
17{,}829 images (35.7\%) are offloaded to the server.

\textbf{Confidence gate.}
The offloading decision follows~\cite{im2024attention} and is controlled by the parameter~$\eta$.
The edge model computes an uncertainty score
$u(x) = -\log_2(\max_c\, p_{\text{local}}(c|x))$, known as min-entropy, and
offloads when $u(x) \geq \eta$.  Higher~$\eta$ leads to fewer offloaded
images (more local processing) but lower local accuracy, creating a
trade-off between communication load and system accuracy.

\textbf{Metrics.}
We report two metrics:
(1)~\emph{Offloaded accuracy}: classification accuracy on images escalated to
the server, which directly measures evidence composition quality;
(2)~\emph{Overall accuracy}: system-level accuracy including both
locally-resolved and offloaded images.

\textbf{Baselines.}
We compare SAGE against five baselines and one reference:
\begin{itemize}
\item \emph{Random Selection}: $B$ patches chosen uniformly at random
  (averaged over 5 seeds).
\item \emph{Uniform Grid}: $B$ patches sampled on a regular spatial grid,
  providing maximum spatial coverage without importance awareness.
\item \emph{Attention Prefix}: top-$B$ patches by descending attention
  score~\cite{im2024attention}---the standard importance-only baseline.
\item \emph{ToMe}~\cite{bolya2023tome}: bipartite soft matching merges the
  most similar token pairs; after merging down to $B$~tokens, the
  representative patch of each group is transmitted.
\item \emph{BAT}~\cite{bat}: decoupled importance--diversity selection that
  splits the budget between attentive and diverse tokens, representing
  the state of the art in importance-plus-diversity token reduction.
\item \emph{Full Transmission}: all 196 patches sent (server ceiling).
\end{itemize}
Each baseline answers a distinct question: Random establishes the lower
bound; Uniform Grid tests whether spatial coverage alone suffices;
Attention Prefix tests whether importance alone suffices; ToMe and BAT
test whether existing importance-plus-diversity methods from the
computational efficiency literature transfer to the communication
setting; and Full Transmission provides the upper bound.

\subsection{Main Results}

Table~\ref{tab:main_results} reports offloaded and overall accuracy
on ImageNet-1K under $\eta\!=\!1.0$.  SAGE achieves the highest accuracy
at every tested budget, outperforming the strongest baseline (BAT) by
$+0.4$--$3.6$\,pp in offloaded accuracy.  The gap is largest at tight
budgets ($+3.6$\,pp at $B\!=\!32$, $+3.4$\,pp at $B\!=\!48$), where
coverage-aware composition matters most.  Notably, Uniform Grid remains
competitive at large budgets ($B\!\geq\!64$), confirming that spatial
coverage alone carries substantial value; yet it falls far behind
content-aware methods at $B\!=\!32$, where importance information is
essential.

These accuracy gains stem from measurably broader spatial coverage.
Table~\ref{tab:coverage_stats} reports $7\!\times\!7$ coarse-grid
coverage, the fraction of spatial grid cells containing at least one
selected patch, across all 17{,}829 offloaded images ($\eta\!=\!1.0$).  SAGE
consistently achieves higher coverage across all budgets, with the gap
peaking at $+3.0$\,pp for $B\!=\!32$.  The advantage is especially
pronounced under hard budgets ($B\!\leq\!64$), confirming that the
importance-then-coverage principle translates directly into reduced
coverage gaps.

\begin{table}[t]
\centering
\caption{Spatial coverage ($7\!\times\!7$ grid, \%) of Attention Prefix
vs.\ SAGE across budgets.  Measured over 17{,}829 offloaded images
($\eta\!=\!1.0$) on ImageNet-1K.}
\label{tab:coverage_stats}
\small
\setlength{\tabcolsep}{5pt}
\begin{tabular}{c|cc|c}
\hline
$B$ & Attn Prefix & SAGE & $\Delta$ \\
\hline
16  & 25.1 & 27.2 & +2.1 \\
32  & 43.0 & 46.0 & +3.0 \\
48  & 56.9 & 59.8 & +2.9 \\
64  & 67.9 & 70.4 & +2.5 \\
80  & 76.8 & 78.7 & +1.9 \\
96  & 83.9 & 85.2 & +1.3 \\
\hline
\end{tabular}
\end{table}

\begin{table*}[t]
\centering
\caption{Main results on ImageNet-1K ($\eta\!=\!1.0$, 17{,}829 offloaded,
50{,}000 total).  Off = offloaded acc.~(\%), Ovr = overall acc.~(\%).
Server ceiling: Off\,=\,64.4\%, Ovr\,=\,80.4\%.
Random: uniform random selection.
Uniform Grid: fixed spatial grid.
Attn Prefix: top-$B$ by attention~\cite{im2024attention}.
ToMe: bipartite token merging adapted for patch selection~\cite{bolya2023tome}.
BAT: importance--diversity decoupled selection~\cite{bat}.}
\label{tab:main_results}
\small
\setlength{\tabcolsep}{4pt}
\begin{tabular}{l|cc|cc|cc|cc|cc|cc}
\hline
 & \multicolumn{2}{c|}{$B\!=\!32$}
 & \multicolumn{2}{c|}{$B\!=\!48$}
 & \multicolumn{2}{c|}{$B\!=\!64$}
 & \multicolumn{2}{c|}{$B\!=\!80$}
 & \multicolumn{2}{c|}{$B\!=\!96$}
 & \multicolumn{2}{c}{$B\!=\!128$} \\
 & Off & Ovr & Off & Ovr & Off & Ovr & Off & Ovr & Off & Ovr & Off & Ovr \\
\hline
Random
  & 10.9 & 61.4 & 24.9 & 66.4 & 39.3 & 71.5 & 48.6 & 74.8 & 54.5 & 76.9 & 60.2 & 78.9 \\
Unif.\ Grid
  & 9.8 & 61.0 & 25.2 & 66.5 & 45.6 & 73.7 & 50.7 & 75.6 & 55.2 & 77.2 & 60.1 & 78.9 \\
Attn Pfx
  & 16.9 & 63.5 & 34.0 & 69.6 & 47.3 & 74.4 & 54.2 & 76.8 & 58.5 & 78.3 & 62.7 & 79.8 \\
ToMe
  & 12.8 & 62.1 & 27.9 & 67.4 & 46.4 & 74.0 & 54.2 & 76.8 & 57.2 & 77.9 & 62.6 & 79.8 \\
BAT
  & 15.6 & 63.0 & 35.0 & 70.0 & 49.0 & 75.0 & 55.9 & 77.4 & 59.8 & 78.8 & 62.5 & 79.8 \\
\textbf{SAGE}
  & \textbf{19.2} & \textbf{64.3}
  & \textbf{38.4} & \textbf{71.2}
  & \textbf{50.2} & \textbf{75.4}
  & \textbf{56.8} & \textbf{77.7}
  & \textbf{60.2} & \textbf{79.0}
  & \textbf{63.1} & \textbf{80.0} \\
\hline
\end{tabular}
\end{table*}

At $B\!=\!96$ with $\eta\!=\!1.0$, SAGE reaches 93\% of the
server ceiling in offloaded accuracy (60.2\% vs.\ 64.4\%) and 98\% in
overall accuracy (79.0\% vs.\ 80.4\%), reducing uplink resource
consumption by more than half.
Fig.~\ref{fig:budget_curve} further shows that SAGE consistently
outperforms Attention Prefix across all budgets and all four
confidence-gate settings ($\eta \in \{0.5, 1.0, 1.5, 2.0\}$); a
detailed analysis of the $\eta$ effect is provided in
Section~\ref{sec:ablation}.

\subsection{Ablation Studies}
\label{sec:ablation}

\textbf{Effect of confidence gate ($\eta$).}
The confidence gate parameter~$\eta$ controls the operating
point of the hybrid system: higher $\eta$ offloads fewer but harder
images.  Fig.~\ref{fig:budget_curve} visualizes the full
budget--accuracy curves across four $\eta$ values, and
Table~\ref{tab:eta_sweep} summarizes the $B\!=\!64$ slice.
Three observations stand out.
First, the accuracy gap widens as the budget decreases: at $B\!=\!32$
the offloaded accuracy gain exceeds $+2$\,pp regardless of $\eta$,
confirming that coverage-aware composition is most valuable when the
budget is tight.
Second, the advantage persists from $\eta\!=\!0.5$, where many images
are offloaded, to $\eta\!=\!2.0$, where only the hardest images are
offloaded, demonstrating that SAGE does not rely on a particular
difficulty profile.
Third, the offloaded-accuracy gap \emph{increases} with $\eta$, from
$+2.6$\,pp at $\eta\!=\!0.5$ to $+3.0$\,pp at $\eta\!=\!2.0$,
indicating that coverage-aware composition is particularly beneficial
for the hardest inputs.

\begin{table}[t]
\centering
\caption{Effect of $\eta$ on accuracy~(\%) at $B\!=\!64$
on ImageNet-1K.  Higher $\eta$ means fewer but harder offloaded images.
Ceiling = server accuracy when all 196 patches are transmitted.
\%Ceil = offloaded accuracy as a fraction of the ceiling.}
\label{tab:eta_sweep}
\small
\setlength{\tabcolsep}{3pt}
\begin{tabular}{c|cc|cc|cc|cc}
\hline
 & \multicolumn{2}{c|}{Ceiling}
 & \multicolumn{2}{c|}{Attn Prefix}
 & \multicolumn{2}{c|}{SAGE}
 & \multicolumn{2}{c}{\%Ceil (Off)} \\
$\eta$ & Off & Ovr & Off & Ovr & Off & Ovr & Attn & SAGE \\
\hline
0.5 & 70.8 & 81.5 & 56.1 & 73.3 & 58.7 & 74.7 & 79.2 & 82.9 \\
1.0 & 64.4 & 80.4 & 47.3 & 74.4 & 50.2 & 75.4 & 73.4 & 78.0 \\
1.5 & 59.6 & 78.1 & 41.1 & 74.2 & 44.1 & 74.8 & 69.0 & 74.0 \\
2.0 & 55.6 & 75.9 & 35.9 & 73.6 & 38.9 & 73.9 & 64.6 & 70.0 \\
\hline
\end{tabular}
\end{table}

\textbf{Prefilter size.}
SAGE's only hyperparameter is the prefilter size (top-$kB$ candidates).
Table~\ref{tab:ablation} reports the effect of varying this
ratio.  At $k\!=\!1$ (i.e., prefilter = budget), SAGE degenerates to
Attention Prefix since there is no room for diversity-based reranking.
At $k\!=\!2$ (the default), sufficient candidates are available for FPS
to find diverse patches while all candidates remain reasonably important.
Larger pools ($k \geq 3$) admit low-attention patches whose noise
offsets their diversity contribution.  No prefilter ($k\!=\!196/B$)
performs worst, confirming that unconstrained diversity without importance
filtering is counterproductive.

\begin{table}[t]
\centering
\caption{Ablation study: prefilter size on ImageNet-1K ($\eta\!=\!1.0$).
Offloaded accuracy~(\%).  Default in \textbf{bold}.}
\label{tab:ablation}
\small
\setlength{\tabcolsep}{3pt}
\begin{tabular}{l|cccccc}
\hline
 & 16 & 32 & 48 & 64 & 80 & 96 \\
\hline
$1B$ (Attn Pfx) & 3.5 & 16.9 & 34.0 & 47.3 & 54.2 & 58.5 \\
\textbf{$2B$ (SAGE)} & \textbf{4.5} & \textbf{19.2} & \textbf{38.4} & \textbf{50.2} & \textbf{56.8} & \textbf{60.2} \\
$3B$             & 3.8 & 16.8 & 35.4 & 48.1 & 55.7 & 59.2 \\
$4B$             & 3.4 & 15.5 & 34.6 & 48.1 & 55.7 & 59.2 \\
196 (no filter)  & 2.9 & 14.7 & 34.4 & 48.1 & 55.7 & 59.2 \\
\hline
\end{tabular}
\end{table}

\subsection{Where Does SAGE Help Most?}

\begin{figure}[t]
\centering
\includegraphics[width=\columnwidth]{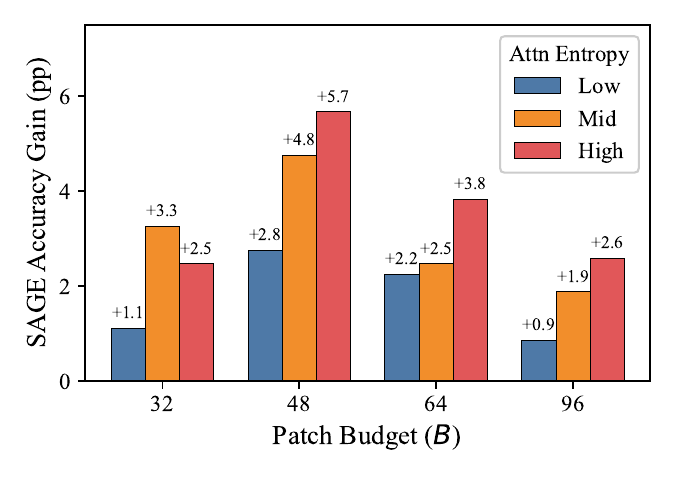}
\caption{SAGE accuracy gain (pp) by client attention entropy tertile
on ImageNet-1K ($\eta\!=\!1.0$).  The high-entropy group benefits most
at every budget, peaking at $+5.7$\,pp for $B\!=\!48$.}
\label{fig:mechanism_depth}
\end{figure}

To identify \emph{which} images benefit most from coverage-aware
composition, we partition all 17{,}829 offloaded images into tertiles by
edge-model \emph{attention entropy}---the Shannon entropy of the
normalized CLS-to-patch attention distribution.  High entropy indicates
that attention is spread across many patches rather than focused on a
few, making top-$B$ selection prone to overlapping coverage.
Fig.~\ref{fig:mechanism_depth} confirms a clear trend: the high-entropy
tertile consistently yields the largest SAGE gain across budgets.
At $B\!=\!48$, the gain rises from $+2.8$\,pp (low entropy) to
$+5.7$\,pp (high entropy)---a $2.0\times$ difference; at $B\!=\!96$ the
ratio is $2.9\times$ ($+0.9 \to +2.6$\,pp).  This pattern
directly supports the coverage mechanism: when client attention is
concentrated, importance-only selection already achieves reasonable
coverage, leaving little room for improvement; when attention is
dispersed, top-$B$ patches cluster around multiple competing regions, and
SAGE's diversity step reclaims the missing coverage.
We also conditioned on edge difficulty (client uncertainty),
but attention entropy was a more reliable predictor of SAGE gain,
suggesting that the \emph{shape} of the attention distribution matters
more than prediction difficulty alone.

\subsection{Backbone Generalization}

A natural concern is whether the coverage gap, and thus SAGE's
benefit, is an artifact of the specific DeiT-Tiny/DeiT-Base pairing.
To test this, we vary the client--server backbone along three
\emph{independent} axes:
(i)~\emph{client capacity}---replacing DeiT-Tiny with DeiT-Small, a
stronger edge model that offloads only the hardest images;
(ii)~\emph{server architecture}---replacing DeiT-Base with
ViT-Base~\cite{dosovitskiy2021vit}, which shares the same architecture
but is trained with a different recipe; and
(iii)~\emph{client architecture family}---replacing DeiT-Tiny with
CaiT-XXS24~\cite{touvron2021cait}, which uses class-attention instead
of standard self-attention.
We evaluate all four pairs on the full ImageNet-1K validation set
(50{,}000 images, $\eta\!=\!1.0$).

\begin{figure}[t]
\centering
\includegraphics[width=\columnwidth]{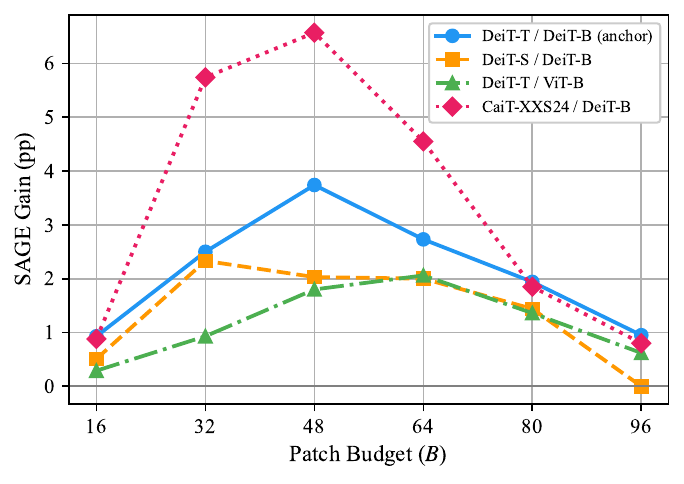}
\caption{SAGE accuracy gain (pp) over Attention Prefix across four
client--server backbone combinations on ImageNet-1K ($\eta\!=\!1.0$).
Despite differences in client capacity, server training recipe, and
attention mechanism, all pairs exhibit the same bell-shaped gain
profile with consistently non-negative gains.}
\label{fig:backbone_generalization}
\end{figure}

As Fig.~\ref{fig:backbone_generalization} shows, all four pairs exhibit
the same qualitative pattern: SAGE gain is positive across all budgets,
peaks in the mid-budget regime ($B\!=\!32$--$48$), and diminishes as the
budget grows.  Across all 24 (pair, budget) combinations, not a single
one shows a negative SAGE gain.
The fact that the same bell-shaped gain profile emerges in every setting
confirms that the coverage gap is a \emph{structural} consequence of
top-$B$ selection, not an artifact of any particular backbone or
training procedure.

%% ====================================================================
\section{Discussion}
\label{sec:discussion}

\textbf{SAGE prioritizes simplicity and generality.}
SAGE uses a fixed prefilter ratio ($2B$) and a single diversity metric
(cosine similarity), requiring no training or additional parameters.
This simplicity is a deliberate design choice: it demonstrates that
even a straightforward instantiation of the importance-then-coverage
principle yields substantial gains.  More sophisticated approaches,
such as learned composition policies, task-conditioned coverage metrics,
or server-feedback-guided composition, may yield further improvements
and represent promising directions for future work.

\begin{figure*}[t]
\centering
\includegraphics[width=0.85\linewidth,trim=0 0 0 8,clip]{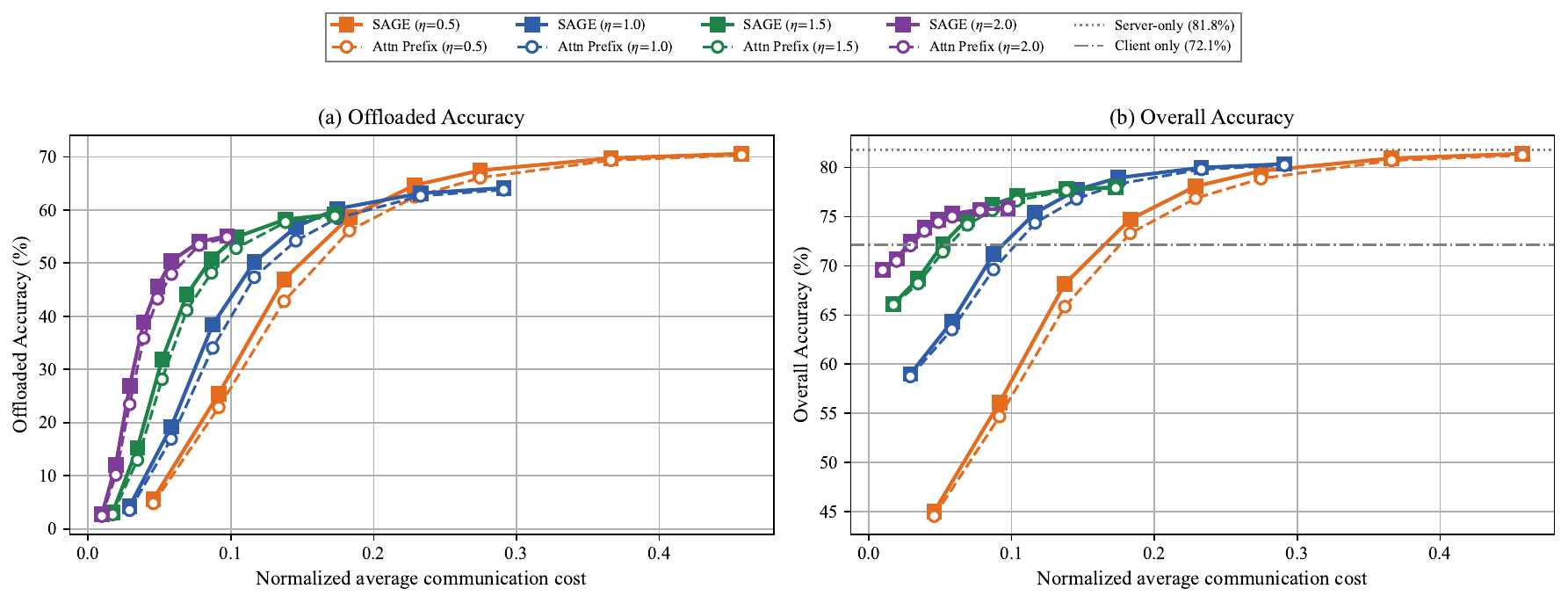}
\caption{Accuracy vs.\ normalized average communication cost on
ImageNet-1K.  Cost is defined as
$\Pr(\text{offload}\mid\eta) \times B / N$, where $N\!=\!196$; a value
of 1.0 corresponds to transmitting all patches for every image.
(a)~Offloaded accuracy and (b)~overall accuracy.  Each curve corresponds
to a fixed $\eta$; points along the curve vary $B$.  SAGE (solid)
consistently outperforms Attention Prefix (dashed) at every operating
point.  Horizontal lines in (b) mark the server-only accuracy (81.8\%) and
client-only accuracy (72.1\%).}
\label{fig:acc_vs_cost}
\end{figure*}

\textbf{System-level communication efficiency.}
Fig.~\ref{fig:acc_vs_cost} replots the results with the x-axis as the
\emph{normalized average communication cost}: the mean number of
transmitted patches per input image divided by the full-transmission
cost ($N\!=\!196$ patches for every image).  A value of 0.05 means the
system uses only 5\% of the communication resources required by
full transmission.  Different $(\eta, B)$ combinations trace out a
cost--accuracy trade-off, and SAGE consistently dominates Attention
Prefix across the entire range.  Notably, at just 5\% of the
full communication cost, SAGE with the optimal operating point
($\eta\!=\!2.0$, $B\!=\!96$) achieves 92\% of the full-transmission
system accuracy (75.3\% vs.\ 81.8\% server-only accuracy).  The figure also reveals
that the optimal confidence threshold $\eta$ depends on the available
communication budget: stringent constraints ($<$10\%) favor high $\eta$
(fewer but better-served offloads), while generous budgets favor low
$\eta$ (more offloads, each with high accuracy).

\begin{figure}[t]
\centering
\includegraphics[width=\linewidth]{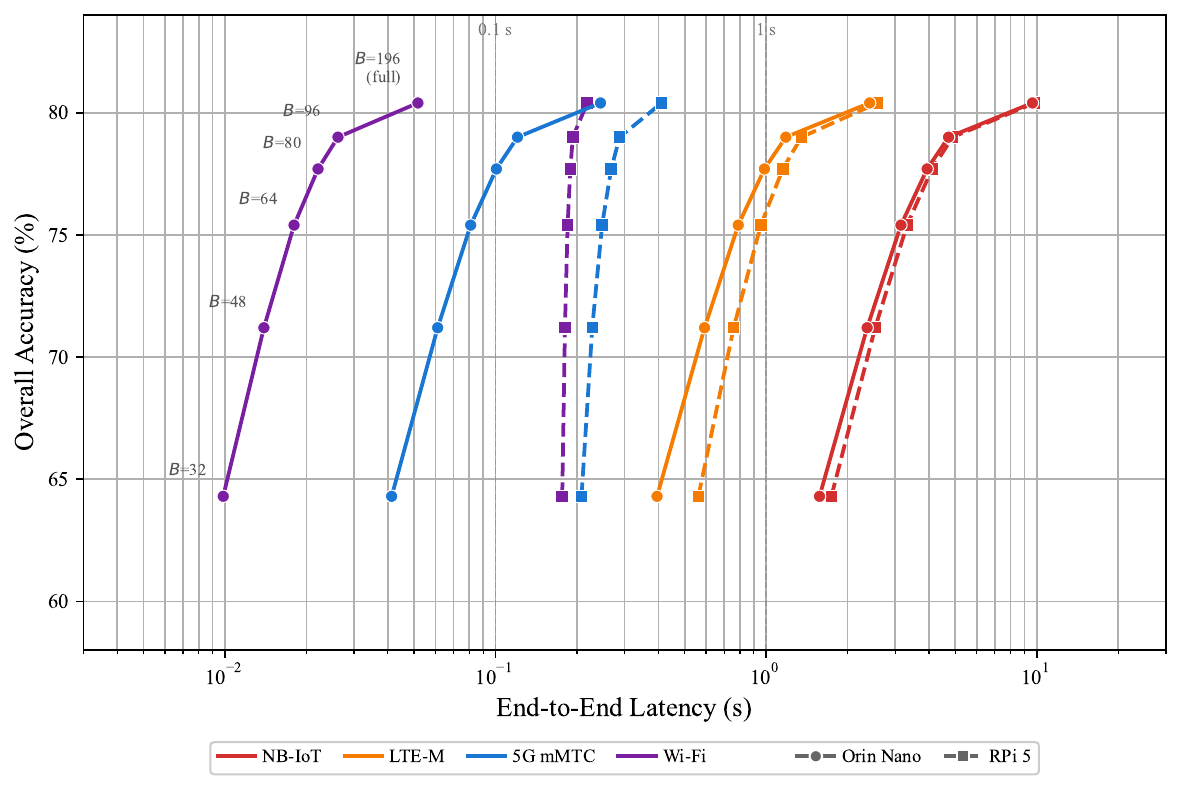}
\caption{Deployment operating points for SAGE.  Each line connects
budgets $B \in \{32, 48, 64, 80, 96, 196\}$ under a fixed
(device, channel) configuration; $B\!=\!196$ corresponds to full
patch transmission (server ceiling 80.4\% overall accuracy).
Colors distinguish uplink technologies; line styles distinguish edge
devices (solid: Orin Nano, dashed: RPi~5; server: NVIDIA~T4 in all
cases).  Vertical dashed lines mark 0.1\,s and 1\,s latency
deadlines.}
\label{fig:operating_points}
\end{figure}

\textbf{Deployment operating points.}
To translate the abstract budget~$B$ into actionable deployment
guidance, Fig.~\ref{fig:operating_points} plots the accuracy--latency
trade-off across 48~operating points spanning six budgets
($B \in \{32, 48, 64, 80, 96, 196\}$), four uplink technologies
(NB-IoT at 250\,kbps, LTE-M at 1\,Mbps, 5G~mMTC at 10\,Mbps, Wi-Fi
at 50\,Mbps), and two edge devices (Jetson Orin Nano at 5\,TFLOPS FP16;
Raspberry~Pi~5 at 30\,GFLOPS), with a shared NVIDIA~T4 inference server.
Each evidence unit is a $16\!\times\!16\!\times\!3$ image patch
transmitted in FP16 ($1.5$\,KB); end-to-end latency sums edge inference,
SAGE selection, uplink transmission, and server inference.
Each line traces a fixed (device, channel) configuration as $B$ increases
from 32 to 196 (full transmission, 80.4\% overall accuracy).
Three observations inform system design.
\emph{First, the uplink channel dominates the operating point.}
For a given accuracy target, end-to-end latency varies by more than two
orders of magnitude across channels: at $B\!=\!96$, SAGE achieves
79.0\% overall accuracy in $\sim$26\,ms over Wi-Fi but requires $\sim$4.7\,s
over NB-IoT.
Reducing $B$ is therefore the single most effective latency lever on
bandwidth-constrained links.
\emph{Second, edge hardware matters only on high-bandwidth links.}
On NB-IoT and LTE-M, the solid (Orin Nano) and dashed (RPi~5) lines
of the same color nearly overlap, because transmission time dwarfs
computation.  On Wi-Fi, the lines separate: RPi~5 requires
$\sim$0.18\,s vs.\ $\sim$0.014\,s on Orin Nano at $B\!=\!48$, making
hardware selection decisive only when the channel is fast enough to
expose compute latency.
\emph{Third, SAGE adds negligible overhead.}
The prefilter step requires $\mathcal{O}(N)$ comparisons and the
iterative FPS loop performs $B$~iterations, each computing cosine
similarities against all $2B$~candidates, yielding an overall
complexity of $\mathcal{O}(N + B^2 D)$ where $D$ is the embedding
dimension.  With $N\!=\!196$, $B\!\leq\!96$, and $D\!=\!192$
(DeiT-Tiny), this amounts to ${\sim}3.5$\,M FLOPs---less than
0.3\% of the edge model's forward pass (${\sim}1.3$\,G FLOPs).
Across all 48~deployment configurations, SAGE patch selection accounts
for $<$0.3\% of end-to-end latency, confirming that the diversity-aware
selection step is practically free.

\textbf{Generality beyond ViT patches.}
Although we evaluate on ViT patch tokens, the coverage gap and the
importance-coverage principle apply to any setting where a
resource-constrained transmitter must compose a limited evidence set from
discrete semantic units: feature-level evidence in split DNNs, multi-modal
token selection, and sensor data fusion in IoT networks.

\textbf{Limitations and future work.}
The current evaluation uses frozen pretrained models.  Server-side
adaptation for sparse evidence inputs could improve absolute accuracy.
Channel-adaptive budgets (where $B$ varies with instantaneous channel
quality), learned composition policies, and extension to video and
multi-modal settings are promising directions.

%% ====================================================================
\section{Conclusion}
\label{sec:conclusion}

We have shown that importance-only evidence composition is
inherently limited in edge--cloud collaborative inference under hard
uplink budgets: the value of an evidence unit lies not in its individual
importance but in its marginal contribution to the information coverage
of the transmitted set.  Server attention analysis confirms that
replacing high-importance units with low-importance but complementary
units improves accuracy, while content-blind spatial coverage alone
achieves competitive performance, demonstrating that coverage is a
first-class design objective alongside importance.  SAGE shows that a
principled, training-free importance-then-coverage method achieves 93\%
of the server ceiling in offloaded accuracy at half the uplink cost,
substantially outperforming importance-only selection without any model
modifications.  The importance-coverage principle identified in this
work opens a new direction for semantic evidence composition in
resource-constrained edge--cloud systems.

\bibliographystyle{IEEEtran}
\bibliography{references}

\end{document}